\documentclass[10pt,twocolumn,letterpaper]{article}

\usepackage{iccv}

\usepackage{times}
\usepackage{epsfig}
\usepackage{subfig}
\usepackage{graphicx} 
\usepackage{amsmath}
\usepackage{amsfonts}
\usepackage{array}
\usepackage{booktabs}
\usepackage{multirow}
\usepackage{makecell}
\usepackage{float} 
\usepackage[dvipsnames]{xcolor}

\usepackage{algorithm}
\usepackage{algpseudocode}

\usepackage[pagebackref=true,breaklinks=true,colorlinks,bookmarks=false]{hyperref}

\usepackage[capitalize]{cleveref}
\crefname{section}{Sec.}{Secs.}
\Crefname{section}{Section}{Sections}
\Crefname{table}{Table}{Tabs.}
\crefname{table}{Table}{Tabs.}

\usepackage{caption}

\iccvfinalcopy 

\def\iccvPaperID{****} 

\ificcvfinal\fi

\begin{document}

\title{Anomaly-Aware Semantic Segmentation via Style-Aligned OoD Augmentation}

\author{Dan Zhang$^{1,2}$, Kaspar Sakmann$^{1}$, William Beluch$^{1}$, Robin Hutmacher$^{1}$, Yumeng Li$^{1,3}$\\
$^{1}$Bosch Center for Artificial Intelligence \  \ 
$^{2}$University of T\"ubingen \ \ 
$^{3}$University of Siegen
\\
{\tt\small \{dan.zhang2,  kaspar.sakmann,  william.beluch,  robin.hutmacher, yumeng.li\}@de.bosch.com
}
}
\maketitle
\ificcvfinal\thispagestyle{empty}\fi

\begin{abstract}
Within the context of autonomous driving, encountering unknown objects becomes inevitable during deployment in the open world. Therefore, it is crucial to equip standard semantic segmentation models with anomaly awareness. Many previous approaches have utilized synthetic out-of-distribution (OoD) data augmentation to tackle this problem. In this work, we advance the OoD synthesis process by reducing the domain gap between the OoD data and driving scenes, effectively mitigating the style difference that might otherwise act as an obvious shortcut during training. Additionally, we propose a simple fine-tuning loss that effectively induces a pre-trained semantic segmentation model to generate a ``none of the given classes" prediction, leveraging per-pixel OoD scores for anomaly segmentation. With minimal fine-tuning effort, our pipeline enables the use of pre-trained models for anomaly segmentation while maintaining the performance on the original task.
\end{abstract}

\section{Introduction}\label{sec:intro}
\begin{figure}%
    \centering
    \subfloat[\centering Confusion Rate]{{\includegraphics[width=4.2cm]{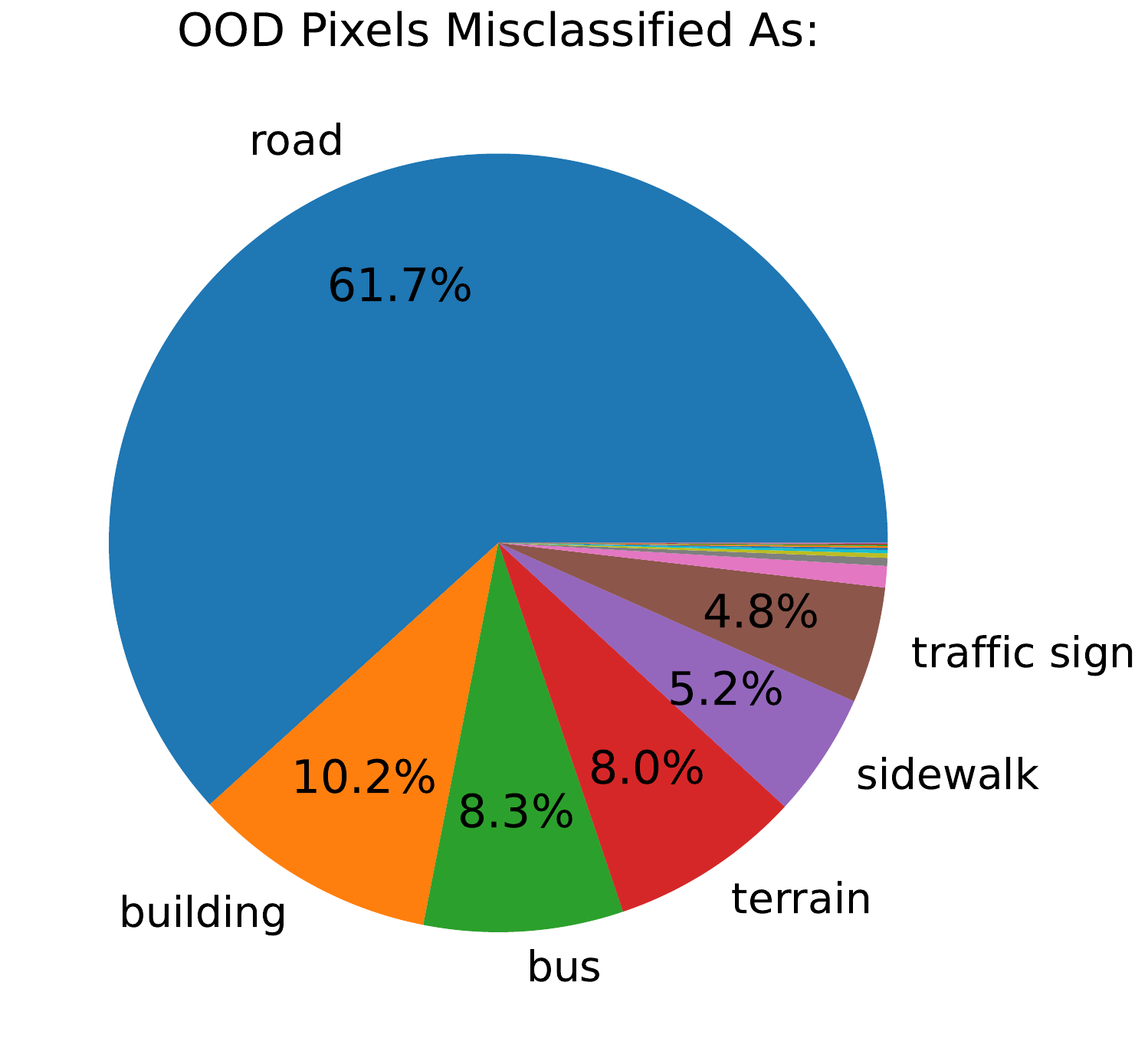} }}%
    \subfloat[\centering Confidence of Error Prediction]{{\includegraphics[width=4.2cm]{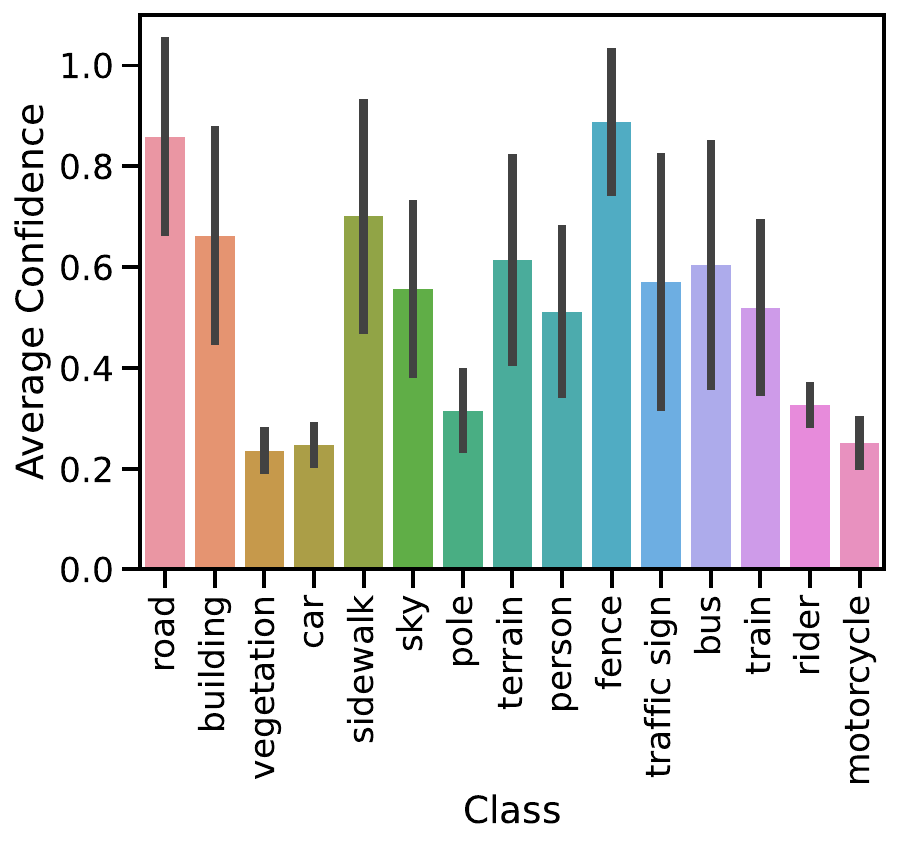} }}%
    \caption{A standard semantic segmentation model, i.e., DeepLabv3+~\cite{chen2017rethinking} with a ResNet101 backbone, is trained on Cityscapes~\cite{Cityscapes} and tested on the anomaly segmentation benchmark Fishyscapes Lost \& Found~\cite{Fishyscapes}. The OoD pixels (anomalies) are frequently misclassified as ``road" with very high confidence. This leads to the highest level of risk for autonomous driving.}%
    \label{fig:teaser}%
\end{figure}

Detecting unknown objects, referred to as Out-of-Distribution (OoD) objects, is vital in safety-critical applications such as  autonomous driving. Anomaly segmentation, combined with standard semantic segmentation, offers precise per-pixel identification of OoD objects in addition to segmenting the in-distribution pixels of the training classes. The main challenge with anomaly segmentation lies in the vast number of potential OoD objects, far exceeding the limited number of training classes. Furthermore, in driving scenes, OoD objects frequently coexist with various known objects, a large difference compared to data in image classification which typically has a single salient object in the center of an image. Class imbalance poses yet another challenge in driving scenes: road pixels often comprise a large part of the training data and bias the network towards predicting road. Due to such imbalances in the training data, networks often exhibit a tendency to predict the majority class for OoD objects found on the road with high confidence, as illustrated in Fig. \ref{fig:teaser}, a critical error that carries a high level of risk.

\begin{figure*}[t!]
    \begin{centering}
    \setlength{\tabcolsep}{0.0em}
    \renewcommand{\arraystretch}{0}
    \par\end{centering}
    \begin{centering}
    \hfill{}
    \begin{tabular}{@{}l@{}c@{}c@{}c@{}c@{}c}
        \centering
         &   &  &  \tabularnewline
        \hspace{0.035\textwidth} Style (Cityscapes) \hspace{0.02\textwidth} \rotatebox{90}{ \hspace{0.015\textwidth} Content \hspace{0.02\textwidth} }

        & {\footnotesize{}}
        \includegraphics[height=0.11\textwidth]{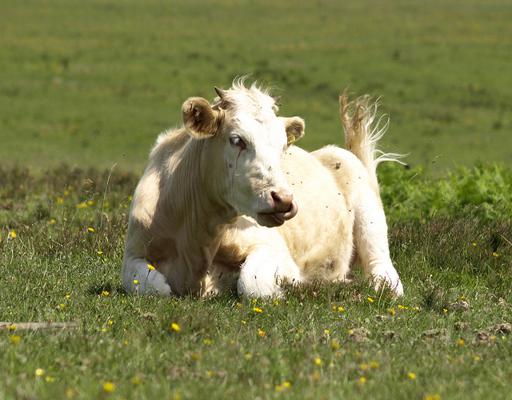} & {\footnotesize{}}
        \includegraphics[height=0.11\textwidth]{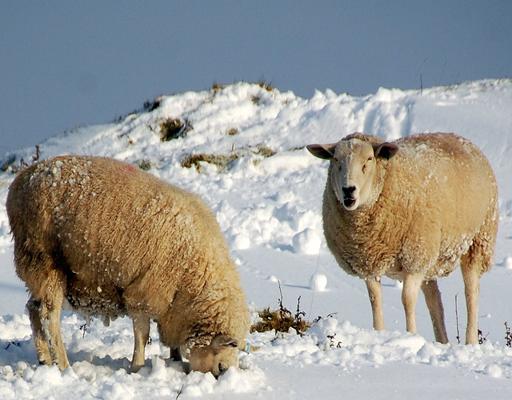} & {\footnotesize{}}
        \includegraphics[height=0.11\textwidth]{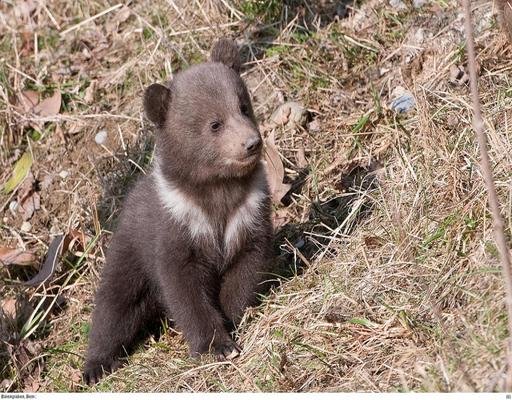} & {\footnotesize{}}
        \includegraphics[height=0.11\textwidth]{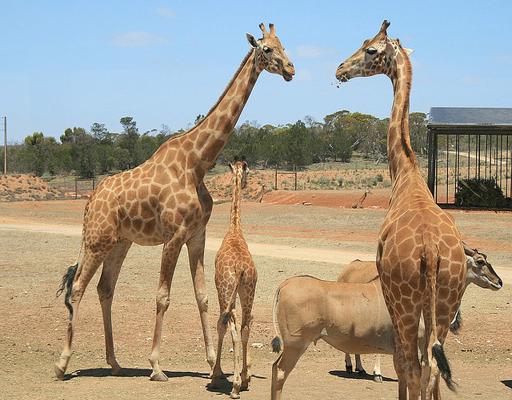}
        \tabularnewline
        
        \includegraphics[height=0.11\textwidth]{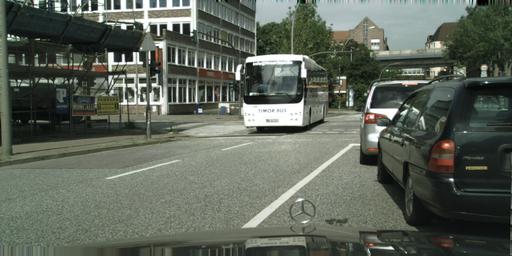} & {\footnotesize{}}
        \includegraphics[height=0.11\textwidth]{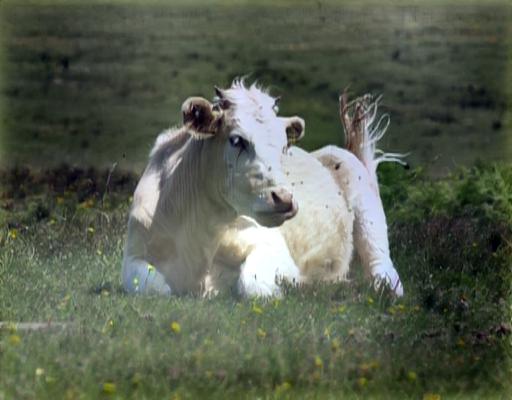} & {\footnotesize{}}
        \includegraphics[height=0.11\textwidth]{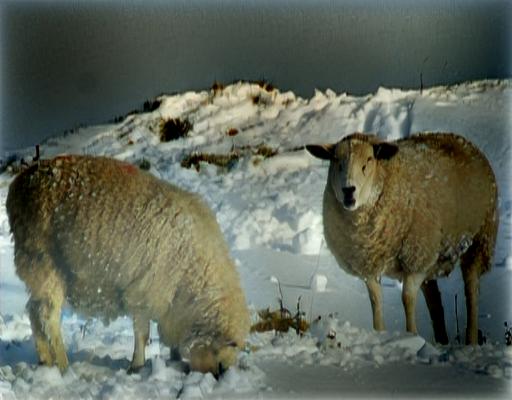} & {\footnotesize{}}
        \includegraphics[height=0.11\textwidth]{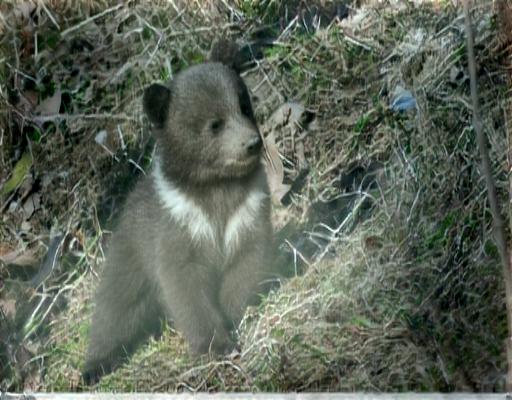} & {\footnotesize{}}
        \includegraphics[height=0.11\textwidth]{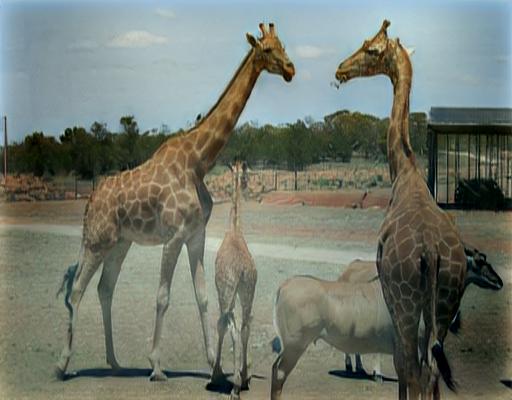}
        \tabularnewline
        
        \includegraphics[height=0.11\textwidth]{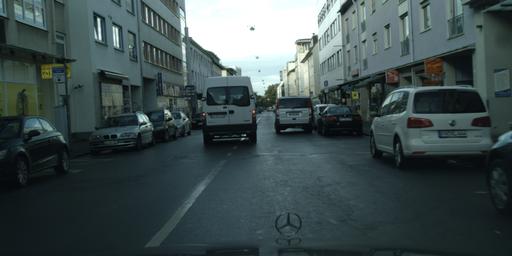} & {\footnotesize{}}
       \includegraphics[height=0.11\textwidth]{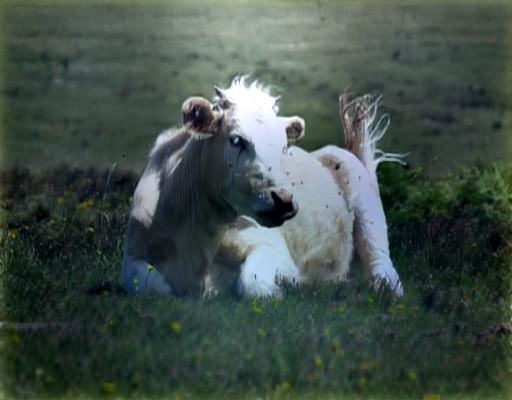} & {\footnotesize{}}
       \includegraphics[height=0.11\textwidth]{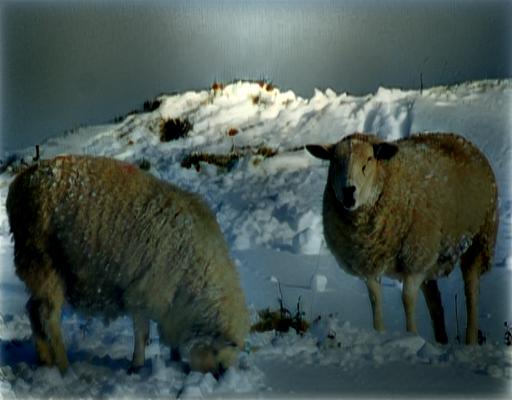} & {\footnotesize{}}
         \includegraphics[height=0.11\textwidth]{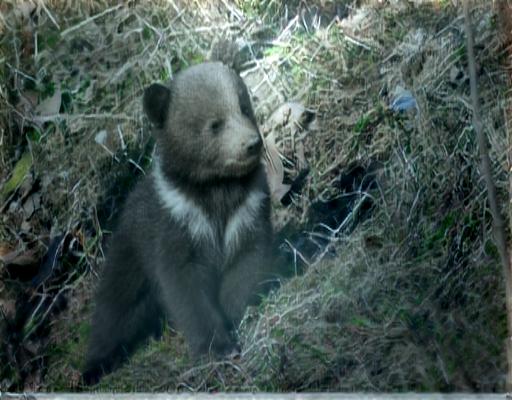} & {\footnotesize{}}
        \includegraphics[height=0.11\textwidth]{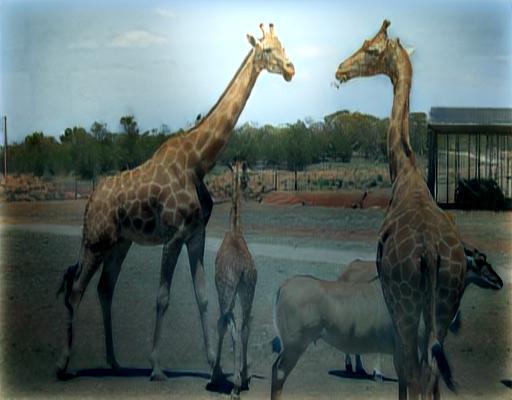}
        \tabularnewline

        \end{tabular}
\hfill{}
\par\end{centering}
\caption{
Visual examples of style transferring from Cityscapes~\cite{Cityscapes} samples to MS COCO~\cite{COCO} objects, using an off-the-shelf ISSA model \cite{li2023intra}. 
}
\label{fig:coco-cs-mix}
\end{figure*}

A particularly successful technique in the literature for anomaly segmentation is a form of outlier exposure \cite{OutlierExposure}. While we do not have access to potential OoD scenarios at inference time, there are many ready-to-use data, different to the training data, which could serve as potential OoD exemplars during training. For instance, MS COCO~\cite{COCO} and ADE20K \cite{Zhou_2017_CVPR} are popular choices for training semantic segmentation on Cityscapes \cite{Cityscapes}, e.g., in \cite{Meta-OOD,SynBoost,PEBAL,liu2022residual, nayal2023rba, grcic22eccv}. %
Earlier work directly added these datasets to the training set \cite{Meta-OOD}, whereas subsequent work discovered that it is more effective to copy and paste objects from them into the training samples of the semantic segmenter, e.g., AnomalyMix in \cite{PEBAL}. Prior works mostly focused on losses and/or architecture changes to incorporate the OoD data during training and fine-tuning, as naively using them can lead to performance degradation \cite{liu2022residual}.

In this work, we discover the benefit of improving the OoD data with simple style transfer. In autonomous driving, the collected training data is often specialized by the camera model, which varies across different vendors. The misalignment between OoD objects from natural images and driving scenes creates a shortcut cue at training, compromising the effectiveness of synthetic OoD data. We thus propose to first align the style before AnomalyMix. Specifically, we utilize ISSA \cite{li2023intra}  to transfer the driving scene data style to OoD objects; see examples in Fig.~\ref{fig:coco-cs-mix}. It was initially proposed to improve domain generalization via style mixing among different training samples.

Equipped with the style-aligned OoD data, we further derive a novel fine-tuning loss to adapt the final classification head of a standard semantic segmentation model for anomaly segmentation, i.e., to output a per-pixel OoD score in addition to the predicted semantic label map. Specifically, we cast per-pixel multi-class classification in semantic segmentation as a set of one vs. the rest (OvR) binary classifications at fine-tuning, i.e., one for each class. The OvR loss allows the prediction of ``none of the given classes", which is the natural ground truth for unknown objects. By considering the top-K losses among the training classes, we specifically optimize for the hard cases and maintain comparably low logit responses from all classes. The loss has a simple form with no hard-to-tune hyperparameters. 

After fine-tuning, we leverage per-pixel OoD scores derived from the logit outputs of the semantic segmenter for anomaly segmentation. Besides the widely used maximum logit \cite{hendrycks2019anomalyseg} and energy-based OoD score \cite{PEBAL}, a new per-pixel OoD score is investigated in this work. It subtracts the maximum logit by the minimum logit. We find such logit difference helpful to tell apart OoD pixels (comparably low logits across all classes) from uncertain in-distribution pixels (e.g., boundary pixels with low maximum logit, but much lower minimum logit). 

Empirically, our proposal delivers impressive improvements on the anomaly segmentation benchmark Fishyscapes Lost \& Found \cite{Fishyscapes}. On another real-world anomaly benchmark Road Anomaly~\cite{RoadAnomaly}, we hypothesize that the domain shift is a key bottleneck when adapting the semantic segmenter pre-trained only on Cityscapes~\cite{Cityscapes}. While unknown objects and domain shifts are often separately tackled in the literature, they can jointly manifest in the real world. This co-occurrence is an interesting yet under-explored topic, awaiting for further investigation.

\begin{figure*}[t!]
\begin{center}
   \includegraphics[width=.95\linewidth]{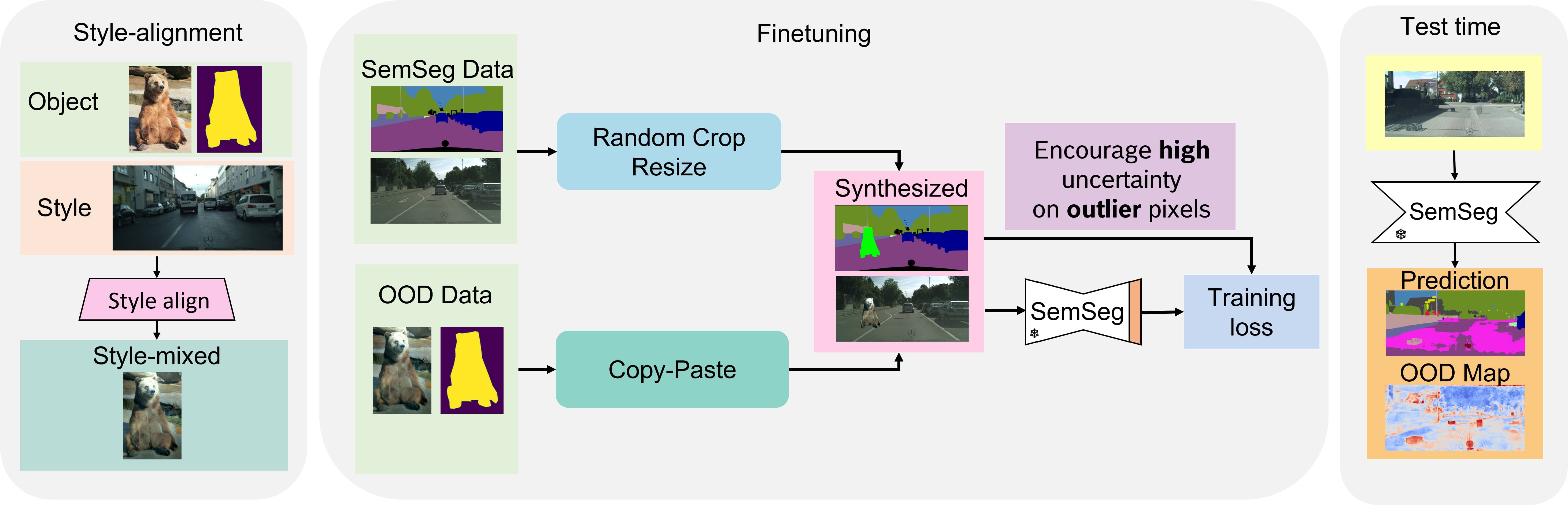}
\end{center}
   \caption{Our proposal begins by aligning the styles between the OoD objects and driving scenes, followed by random copy-pasting. The synthetic OoD data is mixed with the normal driving scenes at a pre-specified ratio. Only the final classification head is fine-tuned; the rest of the network is frozen. At inference, a single-channel OoD map is generated along with the predicted semantic label map. Each pixel of the OoD map carries a scalar score derived from model's output.}
\label{fig:short}
\end{figure*}
\section{Related Work}\label{sec:related_work}

There are two prevalent avenues of approach to tackle anomaly segmentation. The first relies on reconstruction, e.g., \cite{lis2019resynthesis,vojir2021road,xia2020synthesize}, assuming that a poor reconstruction indicates something not well understood by the network. The other avenue adapts semantic segmentation models. Our work aligns with the latter approach. Being different to prior art such as~\cite{besnier2021triggering,liu2022residual}, we attempt to enable anomaly awareness without model architecture changes.

\paragraph{Synthetic OoD Augmentation} Mixing up data samples, despite not necessarily appearing realistic, has proven to be an effective data augmentation strategy, e.g., CutOut~\cite{devries2017cutout}, Mixup~\cite{zhang2018mixup} and CutMix~\cite{yun2019cutmix} for image classification, as well as copy-paste \cite{Copy-Paste} and X-paste~\cite{X-Paste} on object detection and segmentation. 
For the anomaly segmentation task, prior work primarily focused on designing the loss and modifying the architecture to make use of the synthetic OoD data. 
Both~\cite{Meta-OOD} and \cite{PEBAL} adopted a uniform label distribution as the ground-truth label for the OoD pixels, the latter also adding two extra free energy-based regularization terms and the gambler loss \cite{Gamblerloss}. DenseHybrid~\cite{grcic22eccv} also used a free energy-based loss with an additional binary classification head to classify between in-distribution~(ID) and OoD pixels. The authors of \cite{liu2022residual} adopted a contrastive loss, using OoD data as negative examples in contrast to the ID pixels. 
Recent works \cite{nayal2023rba, grcic23cvprw, nekrasov2023ugains, rai2023unmasking} discovered the benefit of using transformer-based models and performing mask level classification for segmentation. Building on top of Mask2Former~\cite{cheng2021mask2former}, the authors of \cite{nayal2023rba, nekrasov2023ugains, rai2023unmasking} used an OvR binary classification loss, while an ensemble over anomaly scores of mask-wide predictions was introduced in~\cite{grcic23cvprw}. 

In this work, we focus on improving the OoD synthesis, while keeping the fine-tuning loss very simple. As the OoD proxy data often comes from a domain very different from driving scenes, the resulting style difference hinders the effectiveness of naive AnomalyMix. With the aid of style-aligned AnomalyMix, a simple variant of a OvR loss already leads to a strong anomaly segmentation performance.

\paragraph{Per-pixel OoD Score} It is a common practice to segment OoD objects based on per-pixel OoD scores, which can be understood as a form of predictive uncertainty on the given training classes. Maximum logit~\cite{hendrycks2019anomalyseg} has been shown as a better choice than maximum Softmax response, and the authors of \cite{StandardizedMaxLogits} improved the latter by correcting the frequent class bias. Free energy was adopted by \cite{PEBAL}, and DenseHybrid \cite{grcic22eccv} further subtracted it by the logit generated by its additional classification head. 

Our fine-tuning method can work with different OoD scores derived from the logits of the semantic segmentation model. Besides the widely used ones (i.e., maximum logit and energy), we additionally introduce a new OoD score, i.e., Max-Min Logit. It computes the difference between the maximum and minimum logit, which helps to better distinguish uncertain ID pixels from OoD pixels, reducing false positive rates (FPRs). 

\section{Method}\label{sec:method}
As shown in Fig.~\ref{fig:short}, our method consists of style-aligned AnomalyMix for fine-tuning a pre-trained model with a top-K OvR loss, and a single-channel OoD map derived from the predictive logits.

\subsection{Style-aligned AnomalyMix}
Similar to AnomalyMix in \cite{PEBAL}, we extract OoD objects from an off-the-shelf dataset, then copy and paste them into the training samples of the semantic segmentation model. Specifically, we only take non-occluded objects from MS COCO and randomly paste them into driving scenes (e.g., Cityscapes) throughout fine-tuning. However, it is quite obvious to see from Fig.~\ref{fig:coco-cs-mix} that MS COCO and Cityscapes have their own domain-specific styles. The network can make use of this style difference to learn anomaly segmentation, which is not a generalizable solution in the real world. Thanks to the advances in generative models, we can transfer the style of training samples to the OoD objects before mixing them together, i.e., aligning the style before AnomalyMix as shown in Fig.~\ref{fig:short}. Technically, we take a pre-trained ISSA model \cite{li2023intra}, which consists of a StyleGAN2-based generator and a masked noise encoder for GAN inversion. The ISSA encoder can respectively extract the style and content of a given exemplar. By mixing up the style and content from different exemplars, the ISSA generator can transfer the style of one exemplar to the other. While the model was pre-trained only on Cityscapes, it successfully transfers the Cityscapes style to unseen MS COCO objects. 

\begin{table*}[t]
\begin{center}
{\small 
    \begin{tabular}{@{}l|c|ccc||c|ccc @{}}
     DeepLabv3+&  \multicolumn{4}{c||}{w. WideResNet38 backbone} & \multicolumn{4}{c}{w. ResNet101 backbone} \\ 
          &  Cityscapes &\multicolumn{3}{c||}{Fishyscapes L \& F} & Cityscapes &\multicolumn{3}{c}{Fishyscapes L \& F} \\ 
        Method & mIoU $\uparrow$ & AUC $\uparrow$ & AP $\uparrow$ &  FPR95 $\downarrow$ & mIoU $\uparrow$  & AUC $\uparrow$ & AP $\uparrow$ & FPR95 $\downarrow$  \\ \midrule
        
        Max Softmax Pred. \cite{hendrycks2019anomalyseg} &  & 89.29 & 4.59 & 40.59  &  & 86.99 & 6.02 & 45.63 \\ 
       
            Max Logit (ML)\cite{hendrycks2019anomalyseg} & & 93.41 & 14.59 & 42.21 & & 92.00 & 18.77 & 38.13  \\
        Entropy \cite{hendrycks2017a}& \textbf{90.62}  & 90.82 & 10.36 & 40.34  &\textbf{80.50} & 88.32 & 13.91&  44.85 \\
        Energy \cite{energy}&  & 93.72 & 16.05 & 41.78   & & 93.50 & 25.79 & 32.26 \\
        Standardized ML \cite{StandardizedMaxLogits} &  & 94.97 & 22.74 & 33.49 &  & 96.88 & 36.55& 14.53  \\
         \midrule
        Meta-OOD \cite{Meta-OOD} & 89.00 & 93.06 & 41.31& 37.69 &  - &  - & - & - \\ 
        PEBAL   \cite{PEBAL} & 89.12 & \textbf{98.96} & 58.81  & \textbf{4.76}   & - &  \textbf{99.09} & 59.83 & 6.49  \\
         \midrule
         \textbf{Ours} (Max Logit)   & & 98.71 & \textbf{71.94}     & 6.42   &  & 98.45 &  67.35 & 9.36 \\
         \textbf{Ours} (Energy)      & 90.39 & 98.79 & {70.87}     & 5.88    & \textbf{80.50}  & 98.58 & \textbf{69.93} & 8.38 \\
         \textbf{Ours} (Max-Min Logit)  &  &  98.87 & {70.84}  & {5.52}    & &  98.83 & 66.32 & \textbf{5.74} 
    \end{tabular}}
\end{center}
\caption{
Comparison of our proposal with prior art on the Fishyscapes Lost \& Found \cite{Fishyscapes}. Note that our comparison scope lies in the line of methods aiming to adapt a standard semantic segmentation model for anomaly segmentation without major architecture changes and extra (sub)networks. The pre-trained model checkpoints and the baseline numbers are taken from the source specified in PEBAL~\cite{PEBAL}. 
}
\label{tab:comparison}
\end{table*}

\subsection{Fine-tuning Loss}
We focus on fine-tuning only the classification head of a pre-trained model, aiming to preserve the integrity of the backbone that is shared with other perception tasks in the autonomous driving system. Under this constraint, grouping the diverse OoD objects into an additional OoD class is suboptimal and can potentially harm the recognition of known objects. While the Softmax-based cross entropy loss over the given training classes is adopted at training, we propose to re-purpose the logit of each class for parameterizing the probability of ``being'' vs. ``not being'' in that class. This effectively turns multi-class classification into a set of binary OvR classifications. On the OoD pixels, all class logits should then be supervised with the ground-truth of ``none of any given classes". The concurrent work~\cite{nayal2023rba} utilized such a loss for both training a Mask2Former based segmenter, and fine-tuning it on synthetic OoD data. Here, we alternatively minimize the top-K OvR losses across the classes
\begin{align}
   \mathcal{L}_{\mathrm{ood}} = \frac{1}{K \vert \mathcal{N}_{\mathrm{ood}}\vert}\sum_{i\in\mathcal{N}_{\mathrm{ood}}} \sum_{k\in \mathcal{S}_{\mathrm{topK}}(i)} -\log \sigma(-s \lambda_{i,k}),
\end{align}
where $\sigma$ is the $\mathrm{sigmoid}$-function, $\mathcal{N}_{\mathrm{ood}}$ is the index set of OoD pixels, $\lambda_{i,k}$ is one of the top-K largest logits $\mathcal{S}_{\mathrm{topK}}(i)$ on the pixel $i$, and $s$ is a hyperparameter to control the slope of the gradients of the $\mathrm{sigmoid}$-function with respect to the logit $\lambda_{i,k}$. Compared to simply averaging the per-class loss over all classes, the top-K variant focuses on improving the worst cases, e.g., the logit of the frequent ``road" class often has a larger response than that of other classes on OoD pixels. Thus, it helps to reduce the logit difference across different classes on the OoD pixels. Besides the loss on the OoD pixels, the total fine-tuning loss is a weighted sum $ \mathcal{L}_{\mathrm{all}} = \mathcal{L}_{\mathrm{id}} + \gamma \mathcal{L}_{\mathrm{ood}}$ with the original training loss on the in-distribution (ID) pixels.

\subsection{Per-pixel OoD score}
After fine-tuning, we resort to per-pixel OoD scores for anomaly segmentation. They are derived from the logits generated by the semantic segmenter. The logit response of an input can be regarded as the negative distance of that sample to the corresponding class, i.e., larger logit value indicates closer to the class prototype (which is the final layer weight vector). In the literature, maximum logit response and negative free energy are top-performing scores. The latter is essentially a smoothed version of the former. Besides using them, we additionally consider the difference between the maximum and minimum logit. In semantic segmentation, there are often many ambiguous in-distribution pixels. For instance, the logit responses of boundary pixels are not as high as that of center pixels in segments. While the maximum logits of boundary pixels may not be sufficiently high, their minimum logits are still much smaller. Therefore, using the gap can help to better tell OoD pixels apart from uncertain in-distribution pixels.   

\section{Experiments}\label{sec:experiment}
In this work, we use semantic segmentation models pre-trained on Cityscapes \cite{Cityscapes}, specifically DeepLabv3+~\cite{chen2017rethinking} with either a Wide ResNet38 or ResNet101 backbone. We use the mIoU ($\%$) as the in-distribution performance metric and compute the area under receiver operating characteristics (AUC), average precision (AP), and the false positive rate at a true positive rate of $95\%$ (FPR95) to validate our approach. The in-distribution dataset is Cityscapes~\cite{Cityscapes}; two real-world anomaly segmentation benchmarks, Fishyscapes Lost \& Found \cite{Fishyscapes} and Road Anomaly \cite{RoadAnomaly} are investigated. For fine-tuning, the default loss configuration is with $\gamma=0.05/0.01$ (WideResNet38/ResNet101), $K=5$, and $s=2$. We adopt the AdamW optimizer with an initial learning rate of $10^{-5}$, polynomial decay at the rate $0.9$, and $20$ fine-tuning epochs in total. The mixing probability for style-aligned AnomalyMix is set as $0.1$. The OoD loss is computed based on the network output before the final bi-linear upsampling operation in DeepLabv3+.

\subsection{Comparison with Prior Art}
The target setting is to adapt an existing semantic segmentation model for anomaly segmentation without architecture changes. Thus, our comparison is with methods that use the same architectures, and can preserve the original mIoU on the Cityscapes validation set. In this case, DeepLabv3+ with the two types of backbones are the most common choices in the literature. As we can see from Table~\ref{tab:comparison}, our method improves over the baselines while preserving the in-distribution performance mIoU. As the methods (in the first block of the table) focused on deriving OoD scores from the pre-trained models, their mIoU stays exactly identical to the original performance. However, such post-hoc OoD score derivation does not lead to strong anomaly segmentation performance, as the pre-trained models are over-confident and uncalibrated as shown in Fig.~\ref{fig:teaser}. Both Meta-OOD~\cite{Meta-OOD} and PEBAL~\cite{PEBAL} trained the model with OoD data, thus outperform the post-hoc methods. Compared to them, we better preserve the original mIoU performance. Notably, on AP, which measures the precision of localizing the OoD pixels, we outperform them by a large margin. Moreover, our method is compatible with different OoD scores. Compared to prior methods that also used Max Logit and Energy, our fine-tuning method greatly improved their performance. Among the three OoD scores, the new score, i.e., Max-Min Logit, is better at reducing FPRs, but slightly worse at increasing APs. It can be an interesting future step to fuse multiple good OoD scores together for further improvements.

\begin{table}[t]
\centering
\resizebox{\columnwidth}{!}{
\begin{tabular}{l|ccc|ccc}
\toprule
& \multicolumn{3}{c}{w./o. Style Align.} & \multicolumn{3}{c}{w. Style Align.} \\
OoD Score &   AUC $\uparrow$  &  AP $\uparrow$  & FPR95 $\downarrow$ &   AUC $\uparrow$ &    AP $\uparrow$ & FPR95 $\downarrow$ \\
\midrule
Max Softmax Pred.   & 94.82 & 32.32 & 20.76 & \textcolor{Green}{$+$1.55} & \textcolor{Green}{$+$18.84} & \textcolor{Green}{$-$2.74} \\
Entropy    & 96.21 & {47.14} & 19.76 & \textcolor{Green}{$+$1.13} & \textcolor{Green}{$+$16.37} & \textcolor{Green}{$-$2.85} \\
Max Logit & 97.84 & 51.79 & 12.61 & \textcolor{Green}{$+$0.61} & \textcolor{Green}{$+$15.56} & \textcolor{Green}{$-$3.25} \\
Energy          & 98.02 & {52.32} & 11.92 & \textcolor{Green}{$+$0.56} & \textcolor{Green}{$+$17.61} & \textcolor{Green}{$-$3.54} \\
Max - Min. Logit & {98.24} & {45.02} &  {9.14} &\textcolor{Green}{$+$0.59} & \textcolor{Green}{$+$21.30} & \textcolor{Green}{$-$3.40} \\
\bottomrule
\end{tabular}%
}

\caption{Style-alignment before AnomalyMix introduces consistent gains across different OoD scores in all metrics. Here, we use DeepLabv3+ (ResNet101) and Fishyscapes Lost \& Found \cite{Fishyscapes}.}
\label{table:2}
\end{table}
\begin{table}[t]
\centering
\resizebox{\columnwidth}{!}{
\begin{tabular}{l|l|ccc}
\toprule
Method & K &    AUC $\uparrow$  &  AP $\uparrow$  & FPR95 $\downarrow$      \\
\midrule
PEBAL~\cite{PEBAL}  & - & \textbf{99.09} &  59.83 & 6.49 \\ \hline
OvR (Max Logit) & - & 97.70 & 52.24 &  12.97 \\
OvR (Energy) & - & 97.95 & 59.96 &  12.09 \\
OvR (Max-Min Logit) & - & 98.52 & 59.19 &   7.51 \\ 
\hline
                          & 3 & 98.34 & 60.86   &  10.50 \\
\textbf{Ours} (Max Logit)                          & 5  & 98.45 & 67.35  &   9.36 \\
                          & 7 & 98.12 & 63.88    &  11.40 \\ \hline
                          & 3 & 98.48 & 64.07 &  9.73  \\
 \textbf{Ours} (Energy)                           & 5 & 98.58 & \textbf{69.93}  &   8.38 \\
                          & 7  & 98.28 & 68.30   &  10.47 \\
                          \hline
                                & 3 & 98.79 & 58.59   &     6.37 \\
\textbf{Ours} (Max-Min Logit)                        & 5   &  {98.83} &   66.32  &  \textbf{5.74} \\
                          & 7 & 98.69 & {66.56}   &   6.43 \\
                          
\bottomrule                          
\end{tabular}
}

\caption{Ablation on the choice of $K$ for our proposal. The fine-tuning loss of PEBAL and OvR involves all classes, whereas only the top-K ones are considered in \textbf{Ours}. Here, we use DeepLabv3+ (ResNet101) and Fishyscapes Lost \& Found \cite{Fishyscapes}.}
\label{table:k}
\end{table}
\begin{figure*}[t]
\begin{center}
   \includegraphics[width=1\linewidth]{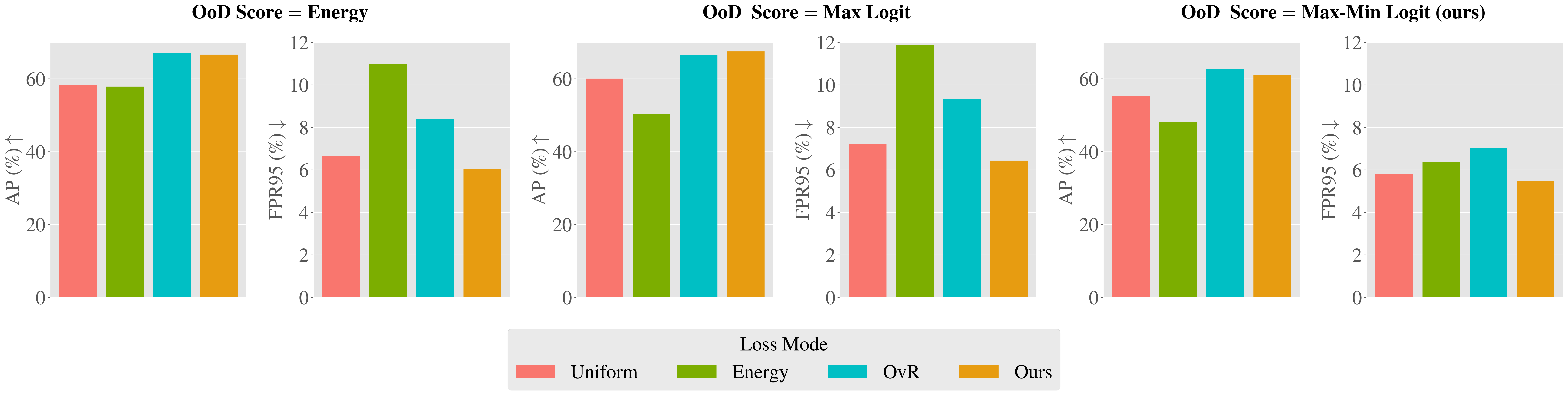}
\end{center}
   \caption{Comparison of different fine-tuning losses, using DeepLabv3+ (ResNet101) and Fishyscapes Lost \& Found. The proposed top-K OvR loss improves the AP and FPR95 when combining with the three top-performing OoD scores.}
\label{fig:onecol}
\end{figure*}
\subsection{Ablation Study}
We first evaluate the benefit from using style alignment before AnomalyMix. As shown in Table.~\ref{table:2}, style alignment is beneficial in all cases, independent of the choice of the OoD score and in all OoD evaluation metrics.

Next, we compare different choices of $K$ for our method in Table~\ref{table:k}. The baseline PEBAL and OvR essentially considers all classes, performing worse than our top-K OvR variant. This reveals the benefit of focusing on improving the worst cases, i.e., the top-K confident predicted classes. Among all three OoD scores, $K=5$ is generally better than $K=3$ and $K=7$, and they all outperform the two competitive baselines. Overall, the performance is insensitive to the value of $K$, making it an easy-to-tune hyperparameter.

Finally, we compare different fine-tuning losses. The two baselines, Uniform and Energy, respectively indicate minimizing the cross entropy with a uniform label distribution, and maximizing the negative Log-Sum-Exp of the multi-class logits. In addition, we compare with the OvR loss averaged over all classes. Compared to them, our top-K OvR loss does not involve all multi-class logits at each optimization step, but rather focuses on the hardest cases, i.e., top-K high logit responses to the OoD pixel. As shown in Fig.~\ref{fig:onecol}, our loss is the only one that excels at both AP and FPR95. Notably, our fine-tuning loss leads to superior performance using all three OoD scores. As all these scores reflect some form of ``uncertainty'' on OoD pixels, the consistent performance improvement indicates our fine-tuning process is effective at inducing a ``none of the given classes'' prediction on anomalies.

\subsection{Visual Results on Fishyscapes Lost \& Found}
\begin{figure*}[t]
\begin{center}
   \includegraphics[width=0.85\linewidth, height=0.68\linewidth]{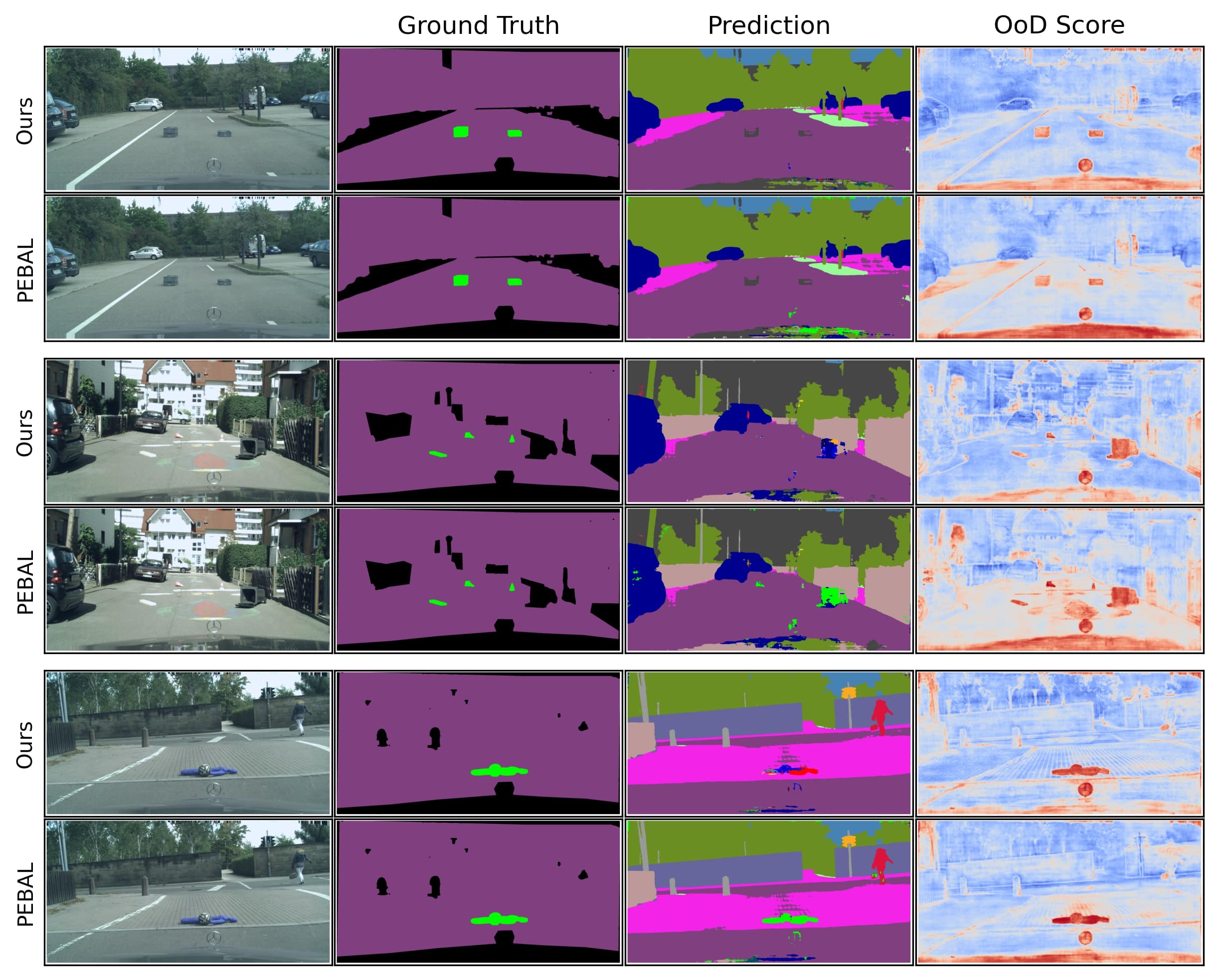}
\end{center}
\vspace{-1em}
   \caption{Visual examples using PEBAL~\cite{PEBAL} and our method for semantic segmentation and anomaly segmentation on Fishyscapes \cite{Fishyscapes}. Our pixel-wise OoD score has less false positive responses to drawings on the road and side walk curbstone, while PEBAL can map some anomalies into its extra class prediction (e.g., dummy person, trash bin), but not always consistently (e.g., boxes and white stick).}
\label{fig:visual}

\end{figure*}

Fig.~\ref{fig:visual} visualizes the semantic label map and per-pixel OoD score produced by PEBAL~\cite{PEBAL} and our method (Max-Min Logit). The training loss of PEBAL has multiple terms. In addition to the uniform label prediction and energy loss, the gambler loss \cite{Gamblerloss} introduced an extra class in prediction. It is quite interesting to observe that some anomalies are classified as the extra class (in green). However, as also pointed out by the authors, it is not yet a reliable prediction for all cases, and the per-pixel OoD score is more informative about the anomalies as we can observe in Fig.~\ref{fig:visual}. Compared to PEBAL, our method uses a single top-K OvR loss without the extra class. The per-pixel OoD score has a strong response to the anomalies, while the false positives, such as paintings on the road or sidewalk, are less frequent than that in PEBAL.

\begin{table}[t]
\begin{center}
{\small 
    \begin{tabular}{@{}l|ccc @{}}
        Method  & AUC $\uparrow$ & AP $\uparrow$ &  FPR95 $\downarrow$   \\ \midrule
        Max Softmax Pred. \cite{hendrycks2019anomalyseg} & 67.53 & 15.72 & 71.38  \\ 
        Max Logit (ML)\cite{hendrycks2019anomalyseg}  & 72.78 & 18.98 & 70.48 \\
        Entropy \cite{hendrycks2017a}  & 68.80 & 16.97 & 71.10  \\
        Energy \cite{energy}  & 73.35 & 19.54 & 70.17  \\
        Standardized ML \cite{StandardizedMaxLogits}  & 75.16 & 17.52 & 70.70  \\
         \midrule
        SynBoost \cite{SynBoost} & 81.91 & 38.21 & 64.75 \\
        PEBAL   \cite{PEBAL} & 87.63 & 45.10 & 44.58   \\
         \midrule
         \textbf{Ours} (Max Logit)  & 89.04 & 43.82 & 39.04 \\
         \textbf{Ours} (Energy)     & \textbf{89.65} & \textbf{46.39} & \textbf{38.09}  \\
         \textbf{Ours} (Max-Min Logit) & 87.15 & 38.38 & 43.40
    \end{tabular}}
\end{center}
\caption{
Comparison of our proposal with prior art on Road Anomaly \cite{RoadAnomaly}, using DeepLabv3+ (WideResNet38). Note that our comparison scope lies in the line of methods aiming to adapt a standard semantic segmentation model for anomaly segmentation without major architecture changes and extra (sub)networks. The pre-trained model checkpoints and the baseline numbers are taken from the source specified in PEBAL~\cite{PEBAL}. 
}
\label{tab:roadanomaly}
\end{table}

\subsection{Domain Shift in Road Anomaly}
\begin{figure*}[t]
\begin{center}
   \includegraphics[width=0.85\linewidth,height=0.94\linewidth]{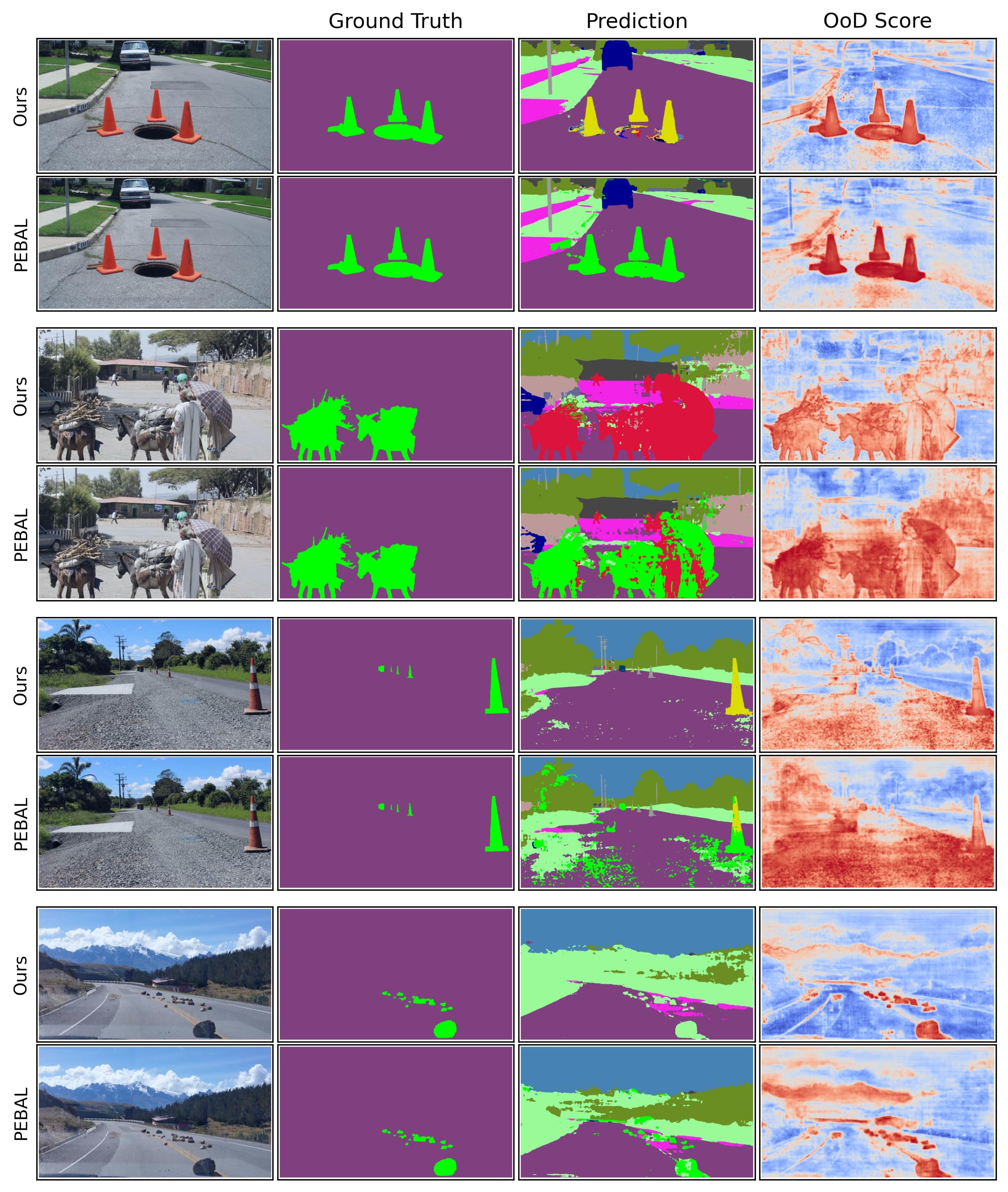}
\end{center}
   \caption{Visual examples using PEBAL~\cite{PEBAL} and our method for semantic segmentation and anomaly segmentation on Road Anomaly \cite{RoadAnomaly}, which is clearly more challenging than Fishyscapes due to the larger domain gap. We can observe a large amount of false positives generated by both methods, where our method is slightly better than PEBAL. It is also worth mentioning that some of them are actually arguable anomalies.}
\label{fig:roadanomaly}
\vspace{-0.5em}
\end{figure*}

Table~\ref{tab:roadanomaly} shows the results on the Road Anomaly \cite{RoadAnomaly} benchmark. While our method still noticeably outperforms the others, all solutions are still far from delivering satisfactory results, with APs below $50\%$ and FPRs above $35 \%$. To gain some insights on the performance gap between Fishyscapes Lost and Found~\cite{Fishyscapes} and Road Anomaly~\cite{RoadAnomaly}, Fig.~\ref{fig:roadanomaly} shows some interesting visual examples. First, we can clearly notice the domain shift from Citypscapes~\cite{Cityscapes}, including not only the style change, but also geographic shift, and object part ambiguity. For instance, the first example is closer to Cityscapes and both approaches handle it well. In the next example, we observe false positives from unfamiliar fences. More interestingly, it is arguable whether the umbrella is an outlier or a part of person in this case. The overall scene is also quite different to the Cityscapes examples, which were collected in Europe. For the last two examples, the rocks on the road should be detected as anomalies, whereas many small stones on the road could be just ``road". To separate both cases, the model will need to have a higher-level concept of the scene if the stone road was not shown during training. Therefore, to tackle the Road Anomaly benchmark, our hypothesis is that focusing on unknown objects with the assumption of no domain shift is suboptimal. It can be interesting to combine anomaly segmentation with domain generalization techniques, e.g., ISSA~\cite{Li_2023_WACV}, RobustNet~\cite{choi2021robustnet}, and StyleLess~\cite{rebut2021styleless}. Moreover, the energy-based OoD score appears to be more robust than Max Logit and Max-Min Logit under the domain shift. Different to the observation on Fishyscapes~\cite{Fishyscapes}, subtracting the minimum logit is no longer effective to reduce false positives. We hypothesize that logits are generally less reliable under the domain shift, and simple logit processing is ineffective to reduce false positives caused by domain shifts. 

Notably, several recent works, i.e.,~\cite{liu2022residual} and~\cite{nayal2023rba,rai2023unmasking,nekrasov2023ugains}, achieved promising progresses, even though they lay beyond the scope of fine-tuning a standard semantic segmentation model. The former added an extra network to Deeplabv3+, and exploited contrastive learning for regularizing the feature space. Contrastive learning is a successful representation learning technique that improves OoD generalization in different tasks. The latter ones built on top of Mask2Former~\cite{cheng2021mask2former}, which is not only transformer-based, but also changes from per-pixel multi-class classification to mask proposals and mask-level classification. Moreover, it is beneficial to pre-train the model on multi-domain driving scene data such as Mapillary~\cite{Mapillary} plus Cityscapes~\cite{Cityscapes}, indicating the value of improved domain generalization achievable by more diverse source data at training.

\section{Conclusion}\label{sec:conclusion}
We proposed a simple fine-tuning process that demonstrates impressive performance gains on the anomaly segmentation benchmark Fishyscapes Lost \& Found. %
The fine-tuned semantic segmentation model preserves the performance on the original task, and additionally generates high-quality per-pixel OoD scores for anomaly segmentation in situations where there is no major domain shift from training/fine-tuning to testing. Notably, our results indicate the value of improving the existing synthetic OoD synthesis process extensively adopted in previous studies. While our proposed style alignment has mitigated the synthetic-to-real gap, it remains an ongoing challenge to completely close the gap. We hope that our findings will inspire further exploration in this direction. Additionally, it is highly interesting to consider the domain shift together with anomalies, e.g., combining with domain generalization techniques for tackling the Road Anomaly challenge.

\clearpage
\clearpage
\bibliographystyle{ieee_fullname}
\bibliography{reference.bib}

\end{document}